# Improving LLM Reliability with RAG in Religious Question-Answering: MufassirQAS

**Ahmet Yusuf Alan**[*1], **Enis Karaarslan**[2], **Omer Aydin**[3,4]

[1] Mugla Sitki Kocman University, Department of Computer Engineering, Türkiye. ahmetyusufalan@posta.mu.edu.tr
[2] Mugla Sitki Kocman University, Department of Computer Engineering, Türkiye. enis.karaarslan@mu.edu.tr
[3] Manisa Celal Bayar University, Department of Electrical and Electronics Engineering, Türkiye, omer.aydin@cbu.edu.tr
[4] University of Waterloo, Management Science and Engineering, Canada, omer.aydin@uwaterloo.ca

**Abstract**

Religious teachings can sometimes be complex and challenging to grasp, but chatbots can serve as effective assistants in this domain. Large Language Model (LLM) based chatbots, powered by Natural Language Processing (NLP), can connect related topics and provide well-supported responses to intricate questions, making them valuable tools for religious education. However, LLMs are prone to hallucinations as they can generate inaccurate or irrelevant information, and these can include sensitive content that could be offensive, inappropriate, or controversial. Addressing such topics without inadvertently promoting hate speech or disrespecting certain beliefs remains a significant challenge. As a solution to these issues, we introduce MufassirQAS, a system that enhances LLM accuracy and transparency using a vector database-driven Retrieval-Augmented Generation (RAG) approach. We built a dataset comprising fundamental books containing Turkish translations and interpretations of Islamic texts. This database is leveraged to answer religious inquiries while ensuring that responses remain reliable and contextually grounded. Our system also presents the relevant dataset sections alongside the LLM-generated answers, reinforcing transparency. We carefully designed system prompts to prevent harmful, offensive, or disrespectful outputs, ensuring that responses align with ethical and respectful discourse. Moreover, MufassirQAS provides supplementary details, such as source page numbers and referenced articles, to enhance credibility. To evaluate its effectiveness, we tested MufassirQAS against ChatGPT with sensitive questions, and our system demonstrated superior performance in maintaining accuracy and reliability. Future work will focus on improving accuracy and refining prompt engineering techniques to further minimize biases and ensure even more reliable responses.

## 1. Introduction

Understanding and learning about religion presents various challenges due to the complexity and depth of religious doctrines and teachings. Accessing accurate and reliable religious information is crucial, especially for individuals who lack formal education in this field. However, a significant risk exists: misinformation that is spread by malicious actors. Throughout history, individuals and groups have exploited people's ignorance and good intentions by distorting religious teachings. This can lead to misunderstandings, misinterpretations, and even harm when individuals rely on unreliable sources for religious knowledge.

To address this issue, there is a need for a system that provides direct, clear, and unambiguous access to trustworthy sources, independent of human intermediaries. Large Language Model (LLM)-based chatbots, which combine artificial intelligence and natural language processing, have emerged as promising solutions in this domain. AI technologies are already widely used across various fields, including natural language understanding and knowledge retrieval [1-8]. Generative AI is widely used in chatbots [9-12], and enables them to generate detailed, context-aware responses to user queries. These systems can process and analyze diverse content, such as text, images, and audio while maintaining a conversational and logical flow. The strength of LLM-based chatbots lies in their ability to link related topics and provide well-informed responses to intricate religious questions, making them well-suited for assisting in religious education.

Today's most widely used chatbots include OpenAI's ChatGPT, Google Gemini, Qwen, DeepSeek, and Anthrophic's Claude [13]. These systems operate on general knowledge databases and generate responses on virtually any topic. While this broad scope is an advantage, LLMs lack domain-specific expertise and may produce incomplete answers. These answers may also be incorrect, which results from a phenomenon known as hallucination. This limitation is particularly problematic in sensitive areas such as religion, where misinformation can have profound consequences.

This study is driven by the following research questions:

- ❖ Can the Retrieval Augmented Generation (RAG) approach help prevent hallucination in LLMs?
- ❖ Can we develop a system that applies this concept to enhance the understanding of Islam?

Our primary contribution is the development of a RAG-based system designed to minimize hallucinations and misinformation in religious inquiries. To achieve this, we constructed a specialized RAG database consisting of the Quran, hadith collections from Kutub-i Sitte, and Islamic catechism texts. This additional retrieval layer is expected to ensure that responses remain accurate and well-supported.

Another key contribution is enhancing chatbot credibility by providing verifiable sources. Traditional chatbots do not typically cite their sources, and this can undermine trust, especially when addressing religious topics that shape individuals' spiritual beliefs. Our system distinguishes itself by explicitly referencing the origin of its responses, reinforcing reliability.

Finally, unlike previous Islamic chatbots that rely solely on literal Quranic translations, our approach incorporates multiple core Islamic sources to ensure more comprehensive and practical answers to everyday religious questions. By integrating hadiths from Kutub-i Sitte and Islamic catechism and the Turkish translation of the Quran, our system generates responses that are contextually informed and accurate.

## 2. Fundamentals

In this section, we will commence by disseminating fundamental information concerning the large language models (LLM) that serve as the primary driving force behind contemporary chatbots, which are progressively aligning themselves with the domain of artificial intelligence. Moreover, we will delve into the vector-based databases in Retrieval Augmented Generation (RAG). Subsequently, we will address the potential challenges encountered in the development of this language model. Additionally, there will be a shared section on the Islamic sources to be used in this study.

### 2.1. Large Language Model and Vector-Based Databases

LLMs are a type of artificial intelligence system that can process and generate natural language texts. They are trained on massive amounts of text data, such as books, articles, web pages, social media posts, and more, using deep neural networks. LLMs can learn the patterns and structures of natural language from the data and use them to perform various tasks, such as answering questions, summarizing texts, translating languages, and writing essays.

Despite their remarkable capabilities, LLMs are not without limitations. Concerns regarding potential biases embedded within training data, factual inaccuracies, and the absence of ethical consideration. Also, they lack explainability and transparency. This makes it nearly impossible to understand how they generate responses and how the LLM works.

### 2.2. Vector Databases and RAG as a Solution

Retrieval-augmented generation (RAG) can be used for knowledge-intensive NLP tasks [14]. RAG is a technique for enhancing the accuracy and reliability of generative AI models with facts fetched from predetermined external sources.

LLMs have some limitations, such as; presenting false information when they do not have the answer, presenting out-of-date or generic information, and creating a response from non-authoritative sources in train data. RAG addresses these challenges by redirecting the LLM to retrieve relevant information from authoritative, predetermined knowledge sources. This way, LLM can use the retrieved data as context for generating a response that is more relevant, accurate, and useful in various contexts. RAG applications potentially provide user transparency by revealing the sources of the retrieved data, offering insight into how the LLM generates its responses.

### 2.3. Islamic Sources

First of all, we can start by learning the word Mufassir and its meaning, which is also mentioned in the title of this study. It is an Arabic word and the name given to theologians who deal with tafsir. Tafsir is mostly used to explain what the Quran's words, compositions and sentences mean. Details about the Tafsir are given below. Now, let's learn the concepts we need to know within the scope of this study, starting from the most basic source of Islam. The Quran is the name of the word that was sent down to the prophet of the Islamic religion through revelation, written on the Mushafs, spread and transmitted through the tongue, and worshipped by reading it [15]. There are different approaches to the root of the name of the Quran. Revelation, on the other hand, is the creator's general informing of the ways of action of beings, and in particular, his transmission to his prophets of all the orders, prohibitions and news that he wants to convey to people, secretly and rapidly, in a direct or unmediated manner [16]. As it is known, the Quran reached the prophet Muhammad through the angel of revelation, Gabriel. It started to be downloaded in the month of Ramadan, one of the Hijri months, and in the Night of al-Qadr [17]. The Quran was transferred from the Preserved Tablet (Levh-i Mahfuz) to the angel of revelation in a manner whose condition we do not know, and through him, it was sent down to the prophet Muhammad at various time intervals [18]. While the Prophet Muhammad was alive, many of the Muslims of the period who believed in him and followed the Prophet's path memorized the whole part of the Quran, so that the Quran was preserved both in writing, on tablets and pages, and in memory by memorization. We should also draw attention to the great efforts made by the Prophet Muhammad in educating the Quran and passing it on to future generations [19].

Since it is impossible to translate the Quran verses into another language without missing any meaning, the translation of the Quran into other languages is called "Meal". This refers to the approximate meaning of the verse. On the other hand, Tafsir (Interpretation) means to declare something, to discover something, to reveal something covered up; It means explaining what is meant by a word whose meaning is unclear and difficult to understand. Tafsir is mostly used to explain what the Quran's words, compositions, and sentences mean. Tafsir is a branch of science. In this branch of science, the Quran is interpreted, and the meanings of words and sentences, their provision,s and wisdom are explained [20]. On the other hand, Hadith is the antonym of "Kadîm". As it is known, Kadim means "Old". Over time, Hadith's meaning is "News" and over time it is created from the infinitive Tahdîs. The word Hadith gained a different meaning in Islam. The words of the Prophet Muhammad are called "el-ehâdîsü'l-kavliyye", his actions are called "el-ehâdîsü'l-fi'liyye", and the things he approves (takrir) are called "el-ehâdîsü't-takrîriyye" [21].

These are the most important resources for people to understand Islam as a faith and to pass it on to the next generations. For this reason, it has become important to learn Islam from the right sources and, at the same time, to convey them using today's technological blessings.

## 3. Related Works

Different use cases for Generative AI and RAG are available in the literature. The use of vector-based databases also finds a place in the literature. LLMs have been shown to experience hallucinations in their responses to instruction-following tasks. Fine-tuning with new data may not prevent hallucinations and may be costly. Therefore, RAG was introduced to provide factual information and adapt to new information [17]. RAG uses a vector database. When a query is made to RAG, the most relevant data from this database is selected. This data is included in the prompt for LLMs. This process can combine existing information without requiring constant fine-tuning and provides context-based information, increasing the clarity and relevance of responses. RAG provides the LLM with the ability to gain a focused understanding vis-à-vis a query. There are studies in the literature where RAG is used in NLP tasks [22-25]. Pal et al. sought to create unique and enjoyable music using Generative AI. When provided with a starting bar of music, the initial Discriminatory network, incorporating Support Vector Machines and Neural Nets, selects a note or chord to guide the subsequent bar. Building upon this chosen chord or note, a Generative Net, which includes Generative Pretrained Transformers (GPT-2) and LSTMs, then generates the entire bar of music. Their innovative two-step approach aims to closely emulate the process of real music composition to enhance the authenticity of the generated music [26].

Kulkarni et al. proposed a Retrieval-Augmented Generator (RAG) approach for constructing a chatbot capable of addressing user queries using Frequently Asked Questions (FAQ) data. The chatbot employs an open API-based paid ChatGPT model as a Language Model (LLM), recognizing the potential to optimize LLM token usage and cost by leveraging previously retrieved context for specific query patterns. The authors employ Reinforcement Learning (RL) to optimize the number of LLM tokens within the constraints of a fixed retrieval model and LLM. The proposed RL-based optimization, coupled with a similarity threshold, achieves notable cost savings with a slight enhancement in accuracy. The study emphasizes the general applicability of the RL approach beyond FAQ chatbots to any existing RAG pipeline [27].

The studies in the literature are classified in Table 1. The study of Mutawa and friends [28] is the oldest study. There are ChatGPT-based studies in the literature such as [29, 30], and online chatbot services such as QuranGPT [31]. Quran and Hadith Datasets [32] study which used an Arabic dataset.

**Table 1.** Classification of the related studies.

| | Al Usouly [28] | QuranGPT [33] | Quran and Hadith Datasets [32] | MufassirQAS |
|---|---|---|---|---|
| **Training Data** | Not clearly specified | Quran | Quran and Hadith (Arabic) | Turkish translation and interpretation of the Quran, Kütüb-i Sitte, Islamic Catechism |
| **Show the source of the answer** | No | No | No | Yes |
| **Hallucination Handling Mechanism** | Not specified | Not specified | Partially by designing the dataset | Yes, with RAG |

In terms of hallucination management, no specific mechanism has been specified in the Al Usouly [28] and QuranGPT [33] systems. The validity of the services based on ChatGPT is questionable as these services do not give references [33]. In the Quran and Hadith Datasets [32] study, partial control has been provided through dataset design, but there is no direct hallucination prevention method. This work mitigates data leakage and memorization by designing a dataset with non-overlapping training and testing sets. MufassirQAS, on the other hand, works with the Retrieval-Augmented Generation (RAG) method, preventing the production of incorrect or out-of-context information and increasing the reliability of the answers.

In terms of training data, the data used in the Al Usouly [28] study is not clearly stated. We are not aware of the training data of the ChatGPT but assume it was trained with the Quran. QuranGPT [33] was trained only with Quran data, while the Quran and Hadith Datasets [32] study used an Arabic dataset containing Quran and Hadith texts, reputable sources, and cross-referencing to ensure full-text inclusion and correctness. MufassirQAS is based on a comprehensive dataset including Turkish Quran translations and interpretations, Kutub-i Sitte, and Islamic Catechism.

In terms of indicating the source of the answers, Other studies do not directly provide the user with the source of the answers, while MufassirQAS provides transparency by providing the relevant sources, page numbers, and articles along with the obtained answers.

When evaluated in general, MufassirQAS offers a more reliable and transparent structure compared to other systems in terms of training data, by its ability to cite sources with a wide and diverse dataset, and its RAG-based hallucination prevention mechanism.

## 4. Method

In this study, we implemented a retrieval-augmented generation (RAG) system to enhance the accuracy and transparency of large language models (LLMs) for natural language question-answering tasks. Rather than relying on unsupervised fine-tuning, we opted for RAG due to its superior performance in handling both pre-existing knowledge encountered during training and entirely new information. As demonstrated by prior research [34], while unsupervised fine-tuning offers some improvement, RAG consistently outperforms it, making it a more effective approach for ensuring accurate responses. We followed the RAG architecture, and the proposed system is shown in Fig. 1.

The study consists of the following components: knowledge base, vector store, question-answering LLM, and user interface. Providing technical details such as data pipeline, retrieval mechanism, and response generation will help to understand the methodology.

The data processing system enters raw input information through multiple sequential stages of the data pipeline. The system begins with data cleaning and normalization tasks, followed by tokenizing the text into specific units. The processing ensures all data sources maintain equivalent formats to prevent inconsistency issues between them. A scalable vector index system processes the preprocessed data for effective retrieval capabilities.

The retrieval module uses a dual method whereby it conducts both semantic retrieval alongside keyword-based retrieval techniques. The system uses the Term frequency-inverse document frequency (TF-IDF) as a keyword method for searching along with transformer embeddings that discover matches with semantic textual coherence between documents.

The response generation step adopts a two-step framework for production. A candidate response set is generated through relevance ranking before going through the neural network re-ranking step. The response generation system verifies both the accuracy and contextual relevance of the results between the processed queries.

The system maintains scalability through superior technical improvements which guarantee both reliability and efficiency.

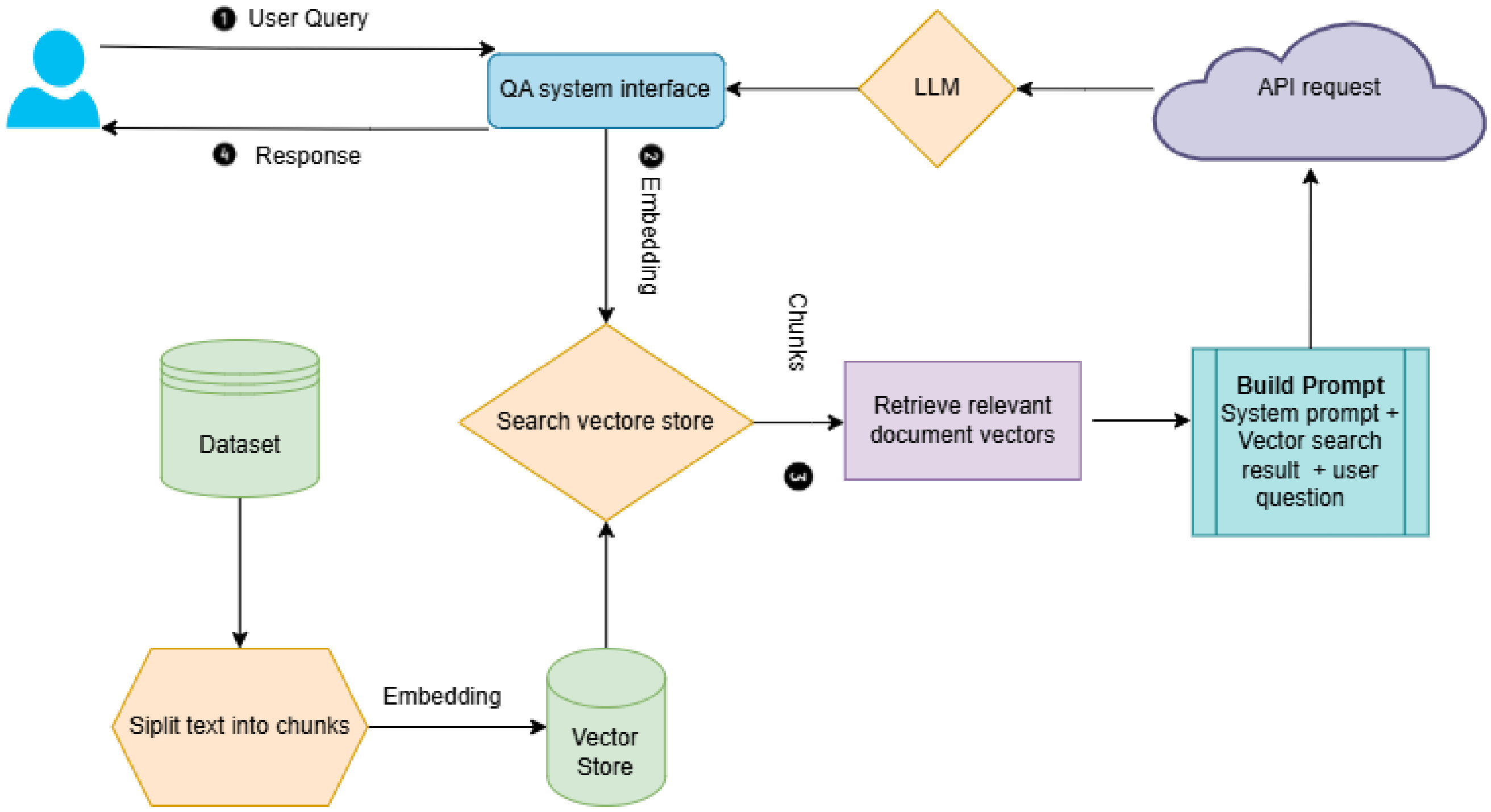

**Fig. 1.** RAG based system

The algorithm of Question answering with the RAG system is shown in Algorithm 1. This algorithm establishes an organized method to use RAG systems for question-answering tasks. It defines the duties of every function to establish a clear understanding of the RAG-based methodology. By combining retrieval and generation, the algorithm effectively enhances the accuracy of responses while maintaining explainability. We will explain the algorithm through specific step-by-step instructions to make each phase understandable. The proposed system is also explained in detail in the subsections:

**Algorithm 1: Algorithm for Question Answering with RAG System**

**Input:** Question $Q$
**Output:** Answer $A$
**Step 1: Create Vector Database**
▷ `Create a vector database from dataset using proper embedding of LLM model`
$D_{\text{enc}} \leftarrow \text{Encode}(D)$
**Step 2: Embed Question**
▷ `Embed Question for vector similarity search`
$Q_{\text{enc}} \leftarrow \text{Embed}(Q)$
**Step 3: Vector Search**
▷ `Rank chunks (document parts) based on relevance to` $Q$ `using vector similarity search on embedded data and retrieve most similar n chunk`
$D_{\text{ranked}} \leftarrow \text{RankChunks}(Q_{\text{enc}}, D_{\text{enc}})$
$R \leftarrow \text{RetrieveChunks}(D_{\text{ranked}})$
**Step 4: Generate Answer**
▷ `Generate answer using LLM with most similar chunks, system prompt and user question`
$A \leftarrow \text{GenerateAnswer}(Q, D_{\text{ranked}})$
**Step 5: Output Answer**
▷ `Display the answer along with relevant document parts for answering the question`
**Output:** $A$

**Algorithm 1.** Algorithm for Question Answering with RAG system

<u>Step 1: Create Vector Database</u>

An appropriate embedding model transforms the dataset $D$ into vector form as the first step. The conversion process allows the data to be accessed through semantic similarity operations with maximum efficiency. The dataset passes through the embedding function Encode(D), which transforms it into the indexed vector database $D_{enc}$. The vector database stores information as the primary knowledge base for all future searches.

<u>Step 2: Embed Question</u>

During user query submission a system embeds $Q$ through Embed(Q) function. Similarity comparisons between $Q_{enc}$ and the encoded dataset become efficient with the produced query vector.

<u>Step 3: Vector Search</u>

The system performs a similarity search between the embedded query $Q_{enc}$ and the encoded document chunks $D_{enc}$ for identifying appropriate information. RankChunks($Q_{enc}$, $D_{enc}$) evaluates the document chunks according to their connection to the query $Q_{enc}$. The most relevant document parts are retrieved through RetrieveChunks($D_{ranked}$) function to choose appropriate contextual information before answer generation.

<u>Step 4: Generate Answer</u>

The system merges identified document portions with the initial user search to create a response. Through the GenerateAnswer($Q$, $D_{ranked}$) function, a language model applies both query and retrieved-context to produce an answer. Model accuracy depends on this step, which gives sufficient details to produce precise contextually proper answers.

<u>Step 5: Output Answer</u>

The system displays the generated answer to the user, including supporting document excerpts as part of transparency measures. The Output(A) function formats the answer and displays it so the users can check the response against the retrieved context.

### 4.1. Forming knowledgebase

We use the PDF format of open-access books to correctly acquire the metadata of books. The following datasets are used to form the vector store:

- ❖ Kuran yolu Türkçe meal ve tefsir (The Way of the Quran - Turkish translation and interpretation) [15, 37]
- ❖ Kütüb-i Sitte [35]
- ❖ İslam İlmihali (Islamic Catechism) [36]

"*Kuran Yolu Türkçe Meal ve Tefsir*" is a prominent scholarly resource that provides a comprehensive Turkish translation and interpretation of the Quran. This work aims to make the Quran accessible to Turkish-speaking audiences while offering deep insights into its verses. It is authored by a team of distinguished Islamic scholars from Türkiye, including Prof. Dr. Hayreddin Karaman, Prof. Dr. Mustafa Çağrıcı, Prof. Dr. İbrahim Kafi Dönmez, and Prof. Dr. Sadrettin Gümüş [37]. This tafsir (exegesis) is noted for its clarity, scholarly rigor, and effort to bridge classical Islamic scholarship with contemporary issues, thus making it a valuable resource for both laypersons and academics interested in Islamic studies.

The Kütüb-i Sitte, which translates to "*The Six Books*," refers to the six most authoritative collections of hadith (sayings and actions of the Prophet Muhammad) in Sunni Islam. These collections are Sahih al-Bukhari, Sahih Muslim, Sunan Abu Dawood, Sunan al-Tirmidhi, Sunan al-Nasa'i, and Sunan Ibn Majah. Together, they play a crucial role in the understanding and practice of Islamic law, theology, and ethics. Each collection was compiled by a different scholar and is revered for its methodological rigor in authenticating the hadiths. The Kütüb-i Sitte is indispensable for scholars of Islamic jurisprudence (fiqh) and hadith studies, providing foundational texts that guide religious practice and legal rulings within the Sunni tradition.

The "*İslam İlmihali*" is an essential resource for Muslims, particularly within the Turkish-speaking community, serving as a comprehensive guide to the fundamentals of Islamic faith and practice. This work is designed to educate Muslims about the core aspects of their religion, covering topics such as belief (aqidah), worship (ibadah), ethics (akhlaq), and social transactions (muamalat). Often used as an instructional manual in religious education, the İlmihal provides clear and practical instructions on performing daily religious duties, understanding Islamic principles, and living an ethical life following Islamic teachings. It is a vital tool for fostering a well-rounded understanding of Islam among believers and is frequently utilized in both formal and informal educational settings.

### 4.2. Creating Vector Store

A vector database is a specific type of database designed to store and retrieve vectors. Vectors are lists of numbers that represent data in a multi-dimensional space. Vector search methods can be used to find similar data by querying for the nearest vectors in the space. This capability is particularly useful for applications that require fast and accurate data matching based on similarity.

To enable vector search, the text in the knowledge base needs to be divided into fixed-length segments called chunks. Since the context of religion issues could be long we assign a higher chunk length. It's important to assign a value to indicate how much overlap each chunk has with the previous one, known as the chunk overlap. To ensure that the language model understands sensitive content and maintains logical connections between parts of the document, we assign a higher value for chunk overlap. Once these values are determined, the chunks should be embedded using the appropriate language model that we utilize for our question-answering system.

### 4.3. Vector Search

After the user submits a question, the system generates an embedding of the user's question. This embedding helps the LLM to comprehend the meaning and context of the query. By using these embedded questions, the LLM can search for relevant parts with the context and assign a similarity score to each part. The system then selects these chunks to form a knowledge context for answering the question. We assign the value of "n" as the number of chunks with the highest similarity score to the question. Considering the abundance of cross-references and diverse explanations surrounding religious matters, we prioritize assigning a higher value to "n." This ensures that the LLM receives a more extensive amount of information regarding the question.

### 4.4. Tuning Prompt

When designing a prompt for an LLM, it is crucial to strategically integrate the system prompt, user queries, and relevant text chunks. The system prompt must clearly define constraints, including the desired response format, tone, and conditions under which the model should abstain from generating a reply. Effective prompt engineering plays a key role in ensuring the chatbot aligns with ethical standards, preventing the production of harmful, offensive, or disrespectful content. Furthermore, the system prompt should be carefully structured to address sensitive topics such as religious beliefs and interfaith conflicts with neutrality and respect. By applying precise prompt engineering techniques, we can enable the chatbot to engage in controversial discussions while avoiding hate speech or content that may offend specific groups.

## 5. Implementation

In our experiments, we employed the ChatGPT-3.5 Turbo version with a 16k context window. We used OpenAI embeddings with a temperature setting of 0.5, and for retrieval-augmented generation (RAG) we relied on the LangChain toolkit [38]. Embedding vectors were stored using Faiss. We configured our processing with a chunk size of 1500 tokens and an overlap of 100 tokens, a setting chosen based on token statistics (see Table 2) and supported by Fig. 2, which shows the token count distribution across various Islamic texts. This configuration was selected to optimize token usage while preserving the complete semantic meaning of each paragraph.

To overcome challenges related to chunk size limitations and context loss, we applied a Recursive Character Text Splitter using a hierarchical method with delimiters in the following order: "\n\n", "\n", and "". This approach first segments the text into larger blocks and then further divides these blocks into smaller, coherent units, thereby enhancing tokenization efficiency. Finally, we presented the top five chunks as the search output. Additionally, Flowise [39] was integrated to display the reference document and vector search results used to answer the questions.

Since we ran the experiments on a local machine and the system utilizes the OpenAI ChatGPT-3.5 model, we didn't need to use high-performance GPU systems. The large language model (LLM) workflow operates on a 12th Generation Intel® Core™ i5-12450H processor with 16GB of RAM.

**Table 2.** Token Statistics of Dataset

| Statistic | Islam Ilmihali | Diyanet Kuran Yolu Turkce meal ve Tefsir | Hadislerle Islam |
|---|---|---|---|
| **Min Tokens** | 27.00 | 37.00 | 5.00 |
| **Max Tokens** | 1585.00 | 1005.00 | 2914.00 |
| **Mean Tokens** | 778.25 | 737.19 | 1022.92 |
| **Median Tokens** | 782.50 | 741.00 | 1155.00 |
| **25th Percentile** | 710.25 | 670.00 | 631.00 |
| **50th Percentile** | 782.50 | 741.00 | 1155.00 |
| **75th Percentile** | 849.50 | 854.00 | 1270.00 |
| **90th Percentile** | 918.00 | 900.00 | 1375.30 |

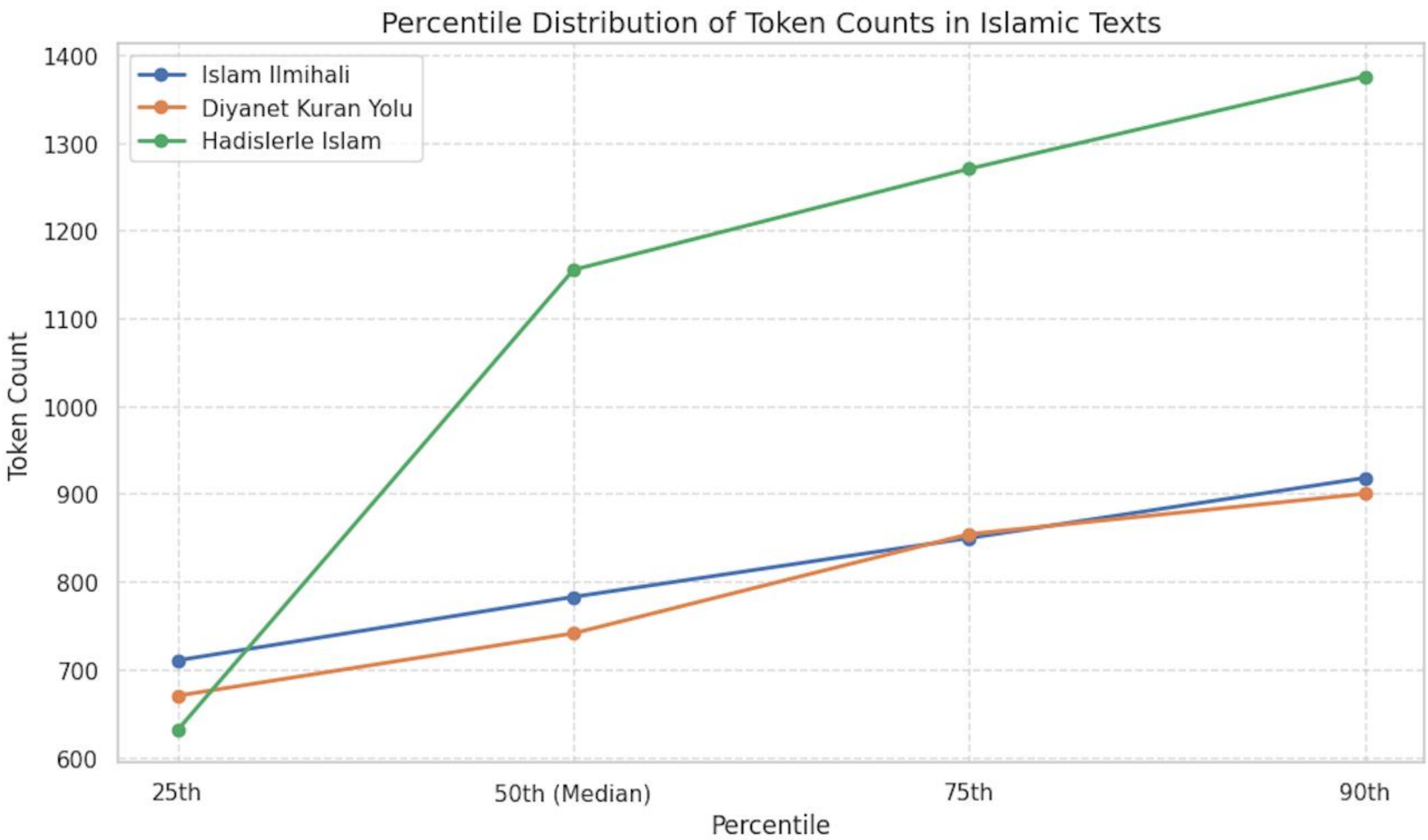

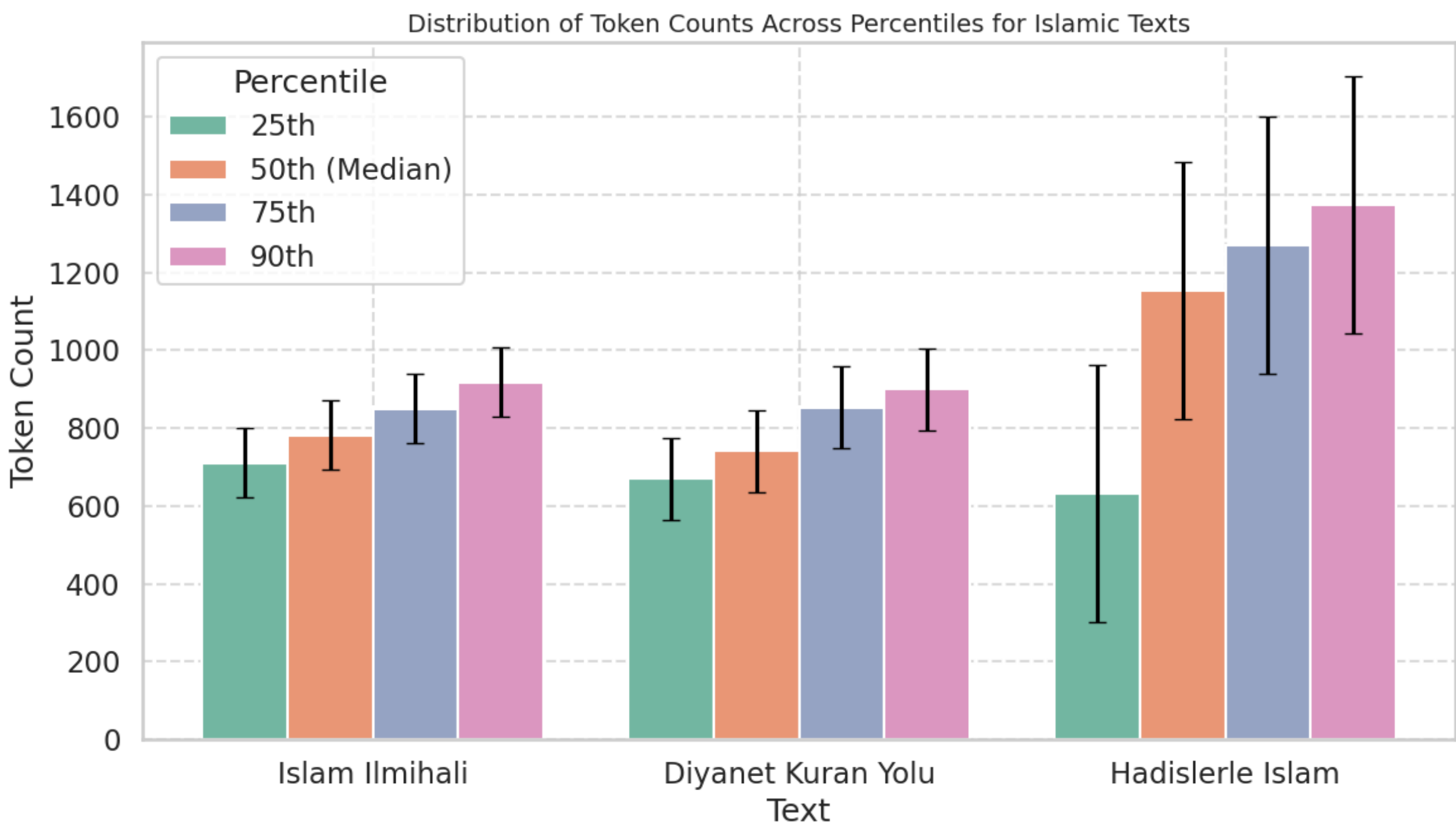

**Fig. 2.** a) Percentile Distribution of Token Counts b) Token Count Range of Islamic Texts

**a)**

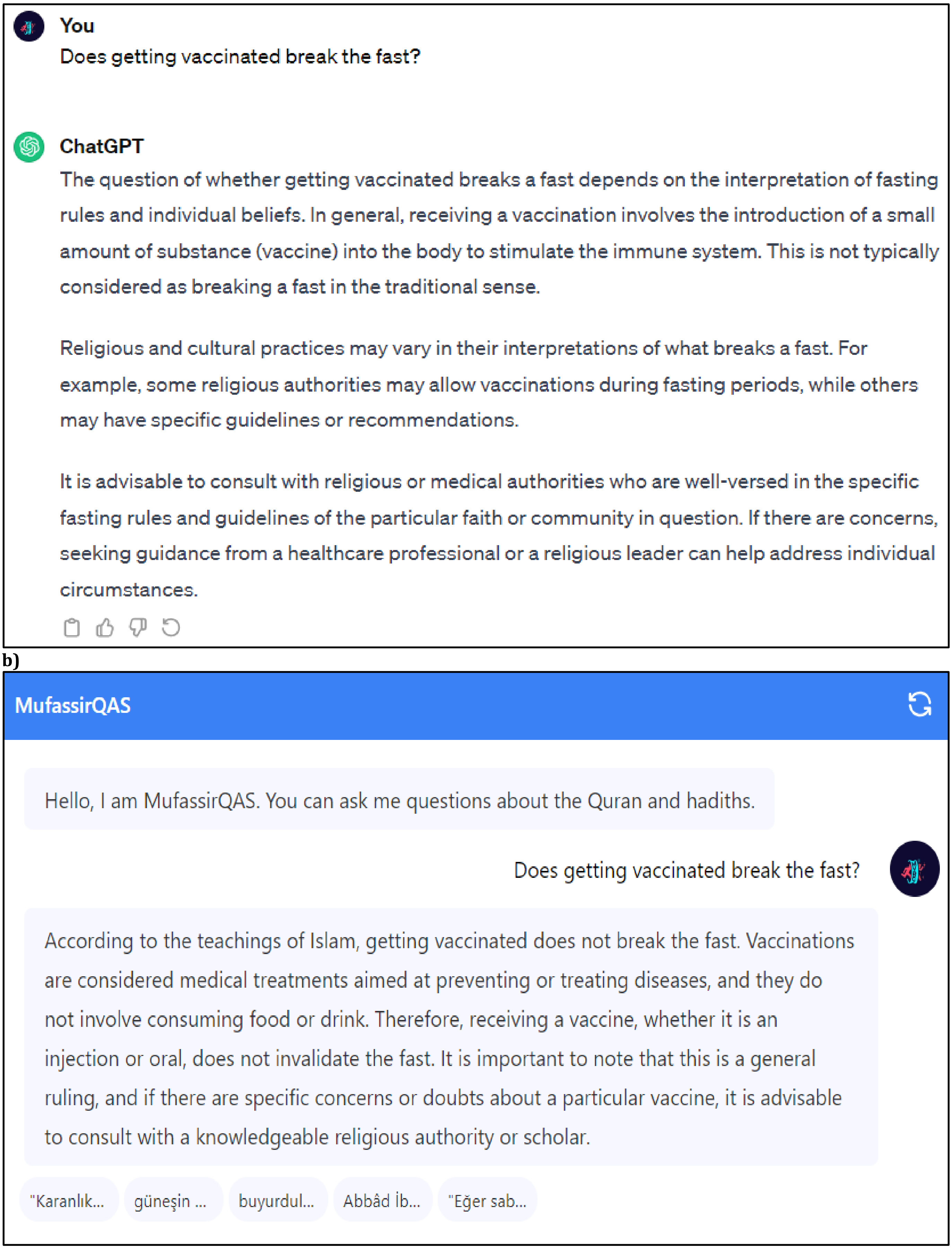

**Fig. 3.** Answer to the Vaccination Question with (a) ChatGpt (b) MufassirQAS

a)

b)

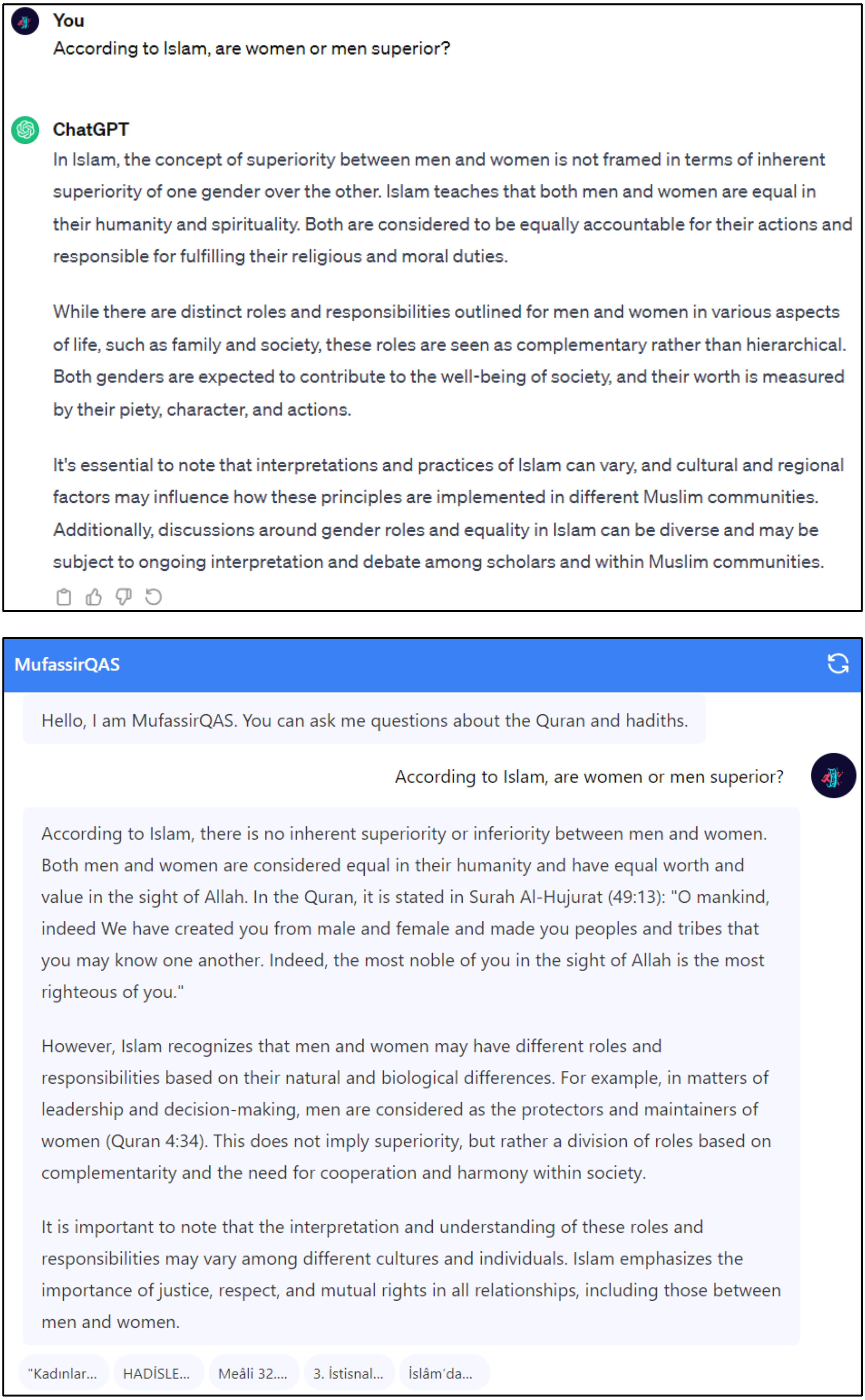

**Fig. 4.** Answer to the Man Woman Equality Question (a) ChatGpt (b) MufassirQAS

```
{
  "pageContent": "Meâli\r\n32. Allah’ın sizi birbirinizden üstün kıldığı şeyleri iç çekerek arzu etmeyin. Erkeklerin de kazandıklarından nasipleri var, kadınların da kazandıklarından nasipleri var. Allah’ın lutfundan isteyin; şüphesiz Allah her şeyi bilmektedir. 33. Ana, baba ve akrabanın geride bıraktıklarından her biri için yakın vârisler belirledik. Antlaşma yoluyla yakınlık bağı kurduğunuz kimselere de paylarını verin. Çünkü Allah her şeyi görmektedir. 34. Allah’ın insanlardan bir kısmını diğerlerine üstün kılmasına bağlı olarak ve mallarından harcama yapmaları sebebiyle erkekler kadınların yöneticisi ve koruyucusudurlar. Sâliha kadınlar Allah’a itaatkârdırlar. Allah’ın korumasına uygun olarak, kimsenin görmediği durumlarda da kendilerini korurlar. (Evlilik hukukuna) baş kaldırmasından endişe ettiğiniz kadınlara öğüt verin, onları yataklarda yalnız bırakın ve onları dövün. Eğer size itaat ederlerse artık onların aleyhine başka bir yol aramayın; çünkü Allah yücedir, büyüktür. 35. Eğer karı-kocanın aralarının açılmasından korkarsanız, erkeğin ailesinden bir hakem ve kadının ailesinden bir hakem gönderin. Düzeltmek isterlerse Allah aralarını bulur; şüphesiz Allah her şeyi bilen, her şeyden haberdar olandır.\r\nTefsiri\r\n32. Hz. Peygamber’in eşi Ümmü Seleme, “Ey Allah’ın elçisi! Erkekler savaşıyor, biz savaşamıyoruz, biz mirasta onların aldığının yarısını alıyoruz” diyerek açıklama istemişti. Yine aynı hanım ve başkaları, kadınların hicretlerinin Kur’an’da zikredilmemesi, bir erkek şahide karşı iki kadın şahidin gerekli bulunmasına paralel olarak amellerin karşılığının da böyle olup olmadığı gibi hususlarda aydınlanma ihtiyacında olduğunu belirtmişti. Bu tür tereddütleri giderici birkaç âyet gelmiş olup bunlardan birisi de açıklamakta olduğumuz âyettir (Tirmizî, “Tefsîr”, 5/8-9).",
  "metadata": {
    "blobType": "",
    "kaynak": "KURAN YOLU TÜRKÇE MEAL VE TEFSİR",
    "loc.lines.from": 2767,
    "loc.lines.to": 2770,
    "source": "blob",
    "yayınevi": "diyanet yayınları",
    "yazar": "Prof. Dr. Hayrettin Karaman , Prof. Dr. İbrahim Kâfi Dönmez , Prof. Dr. Mustafa Çağrıcı , Prof. Dr. Sadrettin Gümüş"
  }
}
```

**Fig. 5.** Example chunk for question “According to Islam, are women or men superior?”

To create the MufassirQAS, we require a platform that combines the mentioned steps and includes a user interface (UI) for interacting with users. This UI should allow users to ask questions, view answers, and relevant information that they use to answer questions. To develop the proof-of-concept (POC) version of this platform, we used a Colab Notebook that utilizes Chroma as a vector database and langchain as the RAG tool provider. However, for the complete platform, we opted to use an existing open-source platform that already possesses the necessary capabilities, rather than building it from scratch. Flowise is one such tool, which enables the creation of a chatbot with RAG capabilities using Langchain. Initially, we hosted Flowise on the Railway cloud host, but due to issues with API calls, we switched to HuggingFace spaces as the cloud-host provider.

Sample screenshots of the implementation are given in Fig. 3 and 4. Turkish versions of the samples are given in the Appendix. The chunk for the question "According to Islam, are women or men superior?" is given in Fig. 5. The answers of the systems are discussed in the next section.

## 6. Results and Discussion

In many general religion-related questions, ChatGPT and MufassirQAS provide similar answers. This is expected because ChatGPT has been trained on a vast amount of data, so it likely includes similar religion-related questions and answers. However, MufassirGPT demonstrates its strength when faced with questions like "Is there evolution according to Islam?" (see Fig. 6), where ChatGPT tends to give uncertain responses and consider the question complex and open to discussion. In contrast, MufassirQAS immediately answers with a clear statement and provides an explanation.

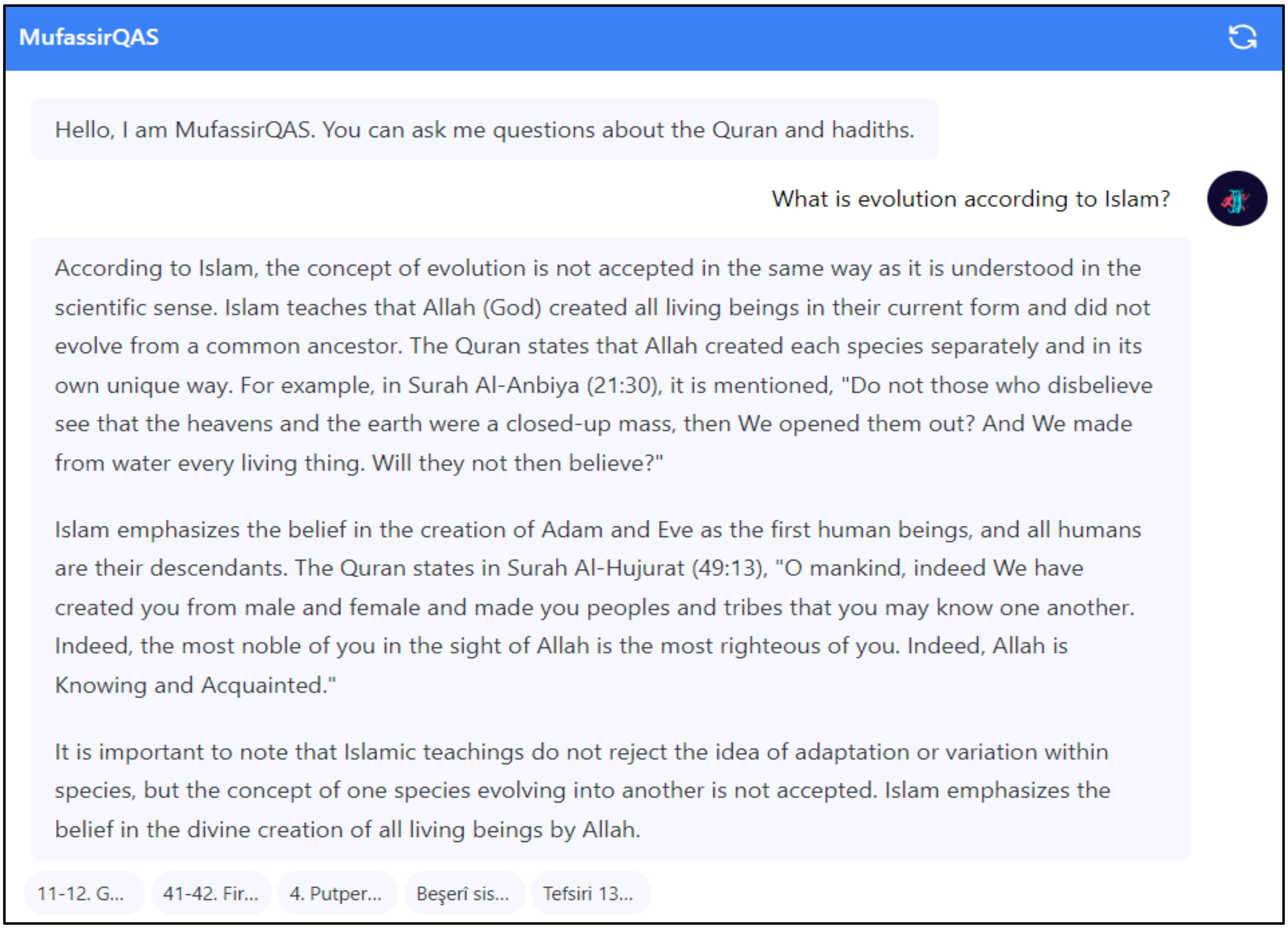

**Fig. 6.** Answer to the Evolution Question

Another distinction between MufassirQAS and ChatGPT is that MufassirQAS tends to provide more references to hadiths and verses, as instructed in the system prompt. For example, when asked "According to Islam, are women or men superior?"; ChatGPT refers to an expression from the Quran without clearly indicating its source (see Fig. 4.a). Meanwhile, MufassirQAS not only shows the users the knowledge source but also directly explains the matter using three Quranic verses (see Fig. 4.b).

There are various interpretations of Islamic issues by different scholars. However, in our study, we adopt an approach based on fundamental sources accepted by all Islamic scholars rather than individual interpretations. The key distinction of the MufassirQAS system is that it directly cites primary sources in its responses. This ensures that the system's answers remain objective and free from personal interpretations.

The sources used in our study consist of works widely accepted in the Islamic world. In the study, primary and reliable sources such as the Turkish translation of the Quran, Kutub-i Sitte, and Islamic Catechism were taken as references. These sources are widely agreed-upon works that reflect the main teachings of Islam. Therefore, the answers given by our system do not prioritize any sectarian or individual view, but only provide information based on the main sources of Islam.

Overall, we can consider MufassirQAS a reliable solution for avoiding vague and unclear responses to questions on Islam. In the study, the Turkish outputs of the results we obtained from MufassirQAS and translated into English are provided in the Appendix of the article. Appendix 1 includes answers to the Vaccination Question in Turkish: (a) ChatGPT, and (b) MufassirQAS. Appendix 2 includes answers to the Man-Woman Equality Question: (a) ChatGPT, and (b) MufassirQAS. Appendix 3 includes answers to the Evolution Question, and Appendix 4 provides an example chunk for the question: "What are the conditions to go to heaven?".

## 7. Limitations of The Study

We acknowledge several limitations, including potential biases in the dataset and the challenge of maintaining religious sensitivity in responses. Additionally, LLMs are prone to security vulnerabilities, which can sometimes result in unexpected outputs. To address these issues, we are continuously refining our approach, working to reduce biases and enhance the system's awareness of religious context to ensure more accurate and respectful responses.

## 8. Conclusion

This study demonstrated that the Retrieval-Augmented Generation (RAG) approach can effectively mitigate hallucinations in LLMs. We developed MufassirQAS, a system that leverages this concept to enhance the understanding of Islam. Through extensive testing with various parameters and questions, we observed that MufassirQAS excels at retrieving relevant information from vector databases and generating accurate responses. It provides well-grounded insights based on the retrieved data.

Interestingly, in many cases, MufassirQAS produced responses similar to those of ChatGPT, despite relying on a significantly smaller dataset. The key distinction is that MufassirQAS tends to provide more definitive answers, whereas ChatGPT often responds with ambiguity and uncertainty. Another notable advantage of MufassirQAS is its ability to cite specific sources from the Quran and hadiths, whereas ChatGPT typically refers to their existence without providing concrete references. Furthermore, we noticed that in some instances, MufassirQAS responses contained more detail than what was explicitly available in the dataset. This occurs because the system utilizes the underlying LLM's (ChatGPT's) general knowledge to bridge informational gaps in the provided data.

However, MufassirQAS has certain limitations. While LLMs are generally proficient at summarizing large texts, setting the chunk size too large can hinder the system's ability to maintain coherence and connect relevant knowledge in its responses. Moreover, when dealing with multiple retrieved chunks, the system sometimes struggles to establish meaningful connections between different pieces of information, which can affect the contextual accuracy of its answers. This issue stems from the system's strict adherence to the retrieved data, occasionally leading to fragmented responses that lack broader contextual understanding.

ChatGPT is a large, well-optimized language model with broad general knowledge. The performance gap between ChatGPT and MufassirQAS will be more pronounced in smaller models that lack extensive training on religious texts and or do not have adequate measures for hallucination reduction. However, there is a limit to how small a model can be while still retrieving and processing relevant information effectively. A major challenge for RAG in smaller models is ensuring that the system correctly understands both the data and user queries to retrieve meaningful information.

As a future work, we aim to explore the scalability of RAG in smaller models, testing the lower bounds of model size while maintaining high retrieval accuracy. Additionally, we plan to develop smaller, more efficient, and sustainable models specialized in religious understanding and question answering, ensuring accessibility and reliability while reducing computational costs.

## Acknowledgement

The study is a part of LLM joint research between MSKU Digital Technologies and Cyber Security (DTCS) Lab and Manisa Celal Bayar University. This paper has been shared as a preprint on Arxiv.org, SSRN.com, and Techrxiv.org. The preprint version can be reached at https://dx.doi.org/10.2139/ssrn.4707470, https://doi.org/10.48550/arXiv.2401.15378 or https://doi.org/10.36227/techrxiv.170723304.41988020/v2

## Author contributions

All authors participated the preparing the manuscript. A.Y.A. conducted the analyses related to the RAG database and contributed to drafting the results. E.K. and O.A. led the primary writing of the manuscript and provided scientific insights that shaped the development of the article. O.A. assisted A.Y.A. with the RAG database work and experimental procedures, as well as contributing to the manuscript drafting. All authors approved this final published version of the manuscript.

## Conflicts of interest

The authors certify that there is no conflict of interest with any financial organization regarding the material discussed in the manuscript.

## Funding

This research received no specific grant from any funding agency in the public, commercial, or not-for-profit sectors.

**Data Availability**

The books, articles, and documents used in this study are open access. If necessary, data can be obtained from the internet or by request from the authors.

**Appendix.** Turkish Versions of The Samples

**a)**

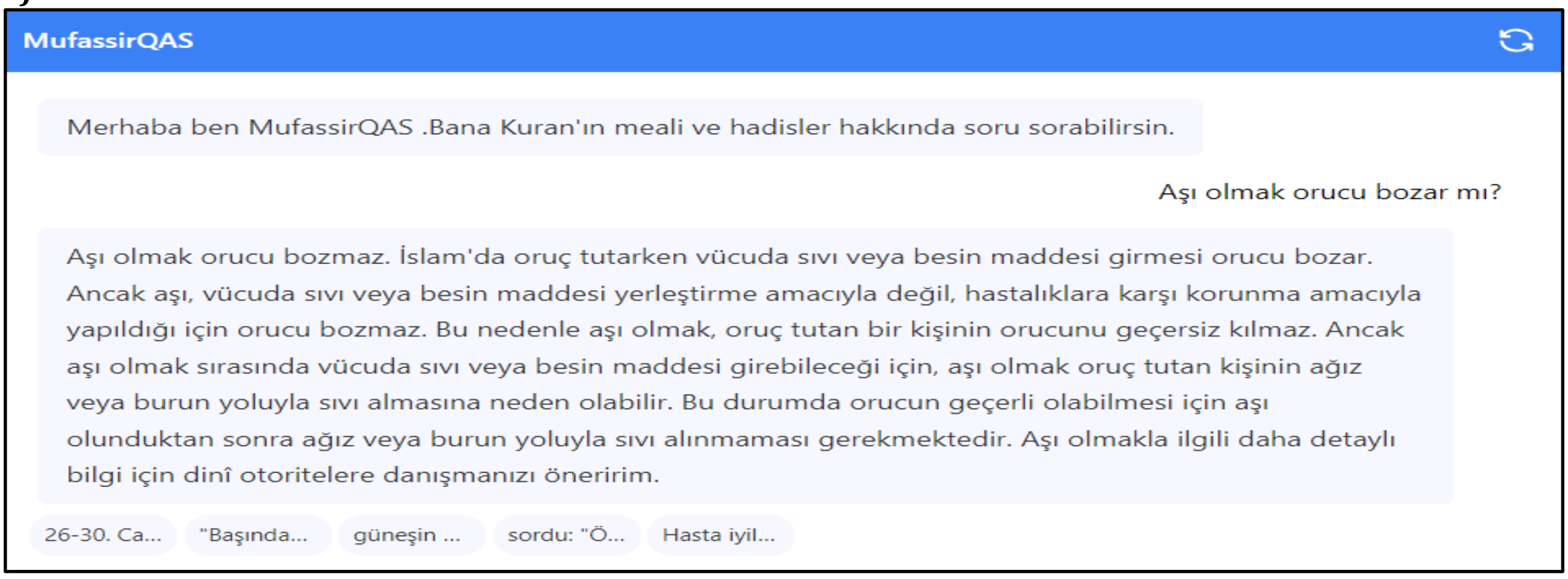

**b)**

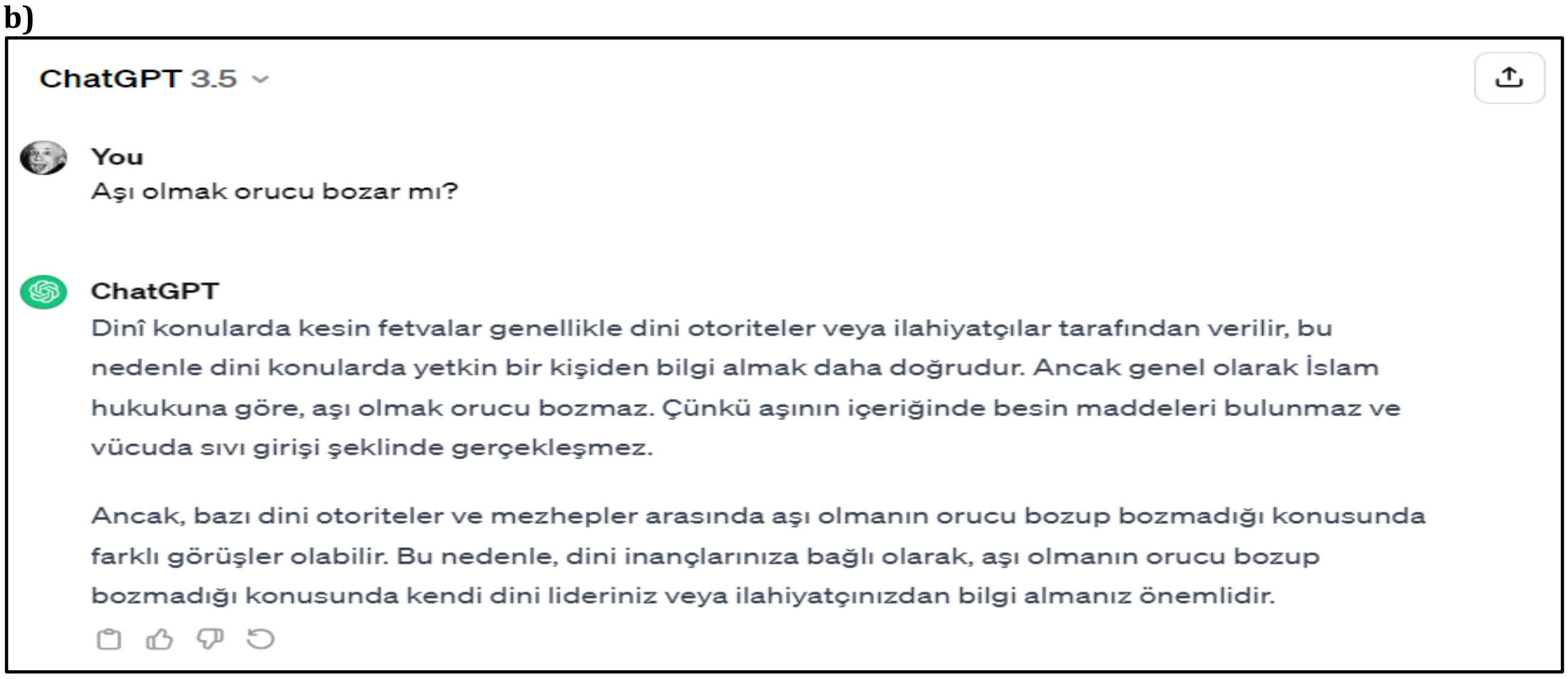

**Appendix 1.** Answer to the Vaccination Question in Turkish (a) ChatGPT (b) MufassirQAS**a)**

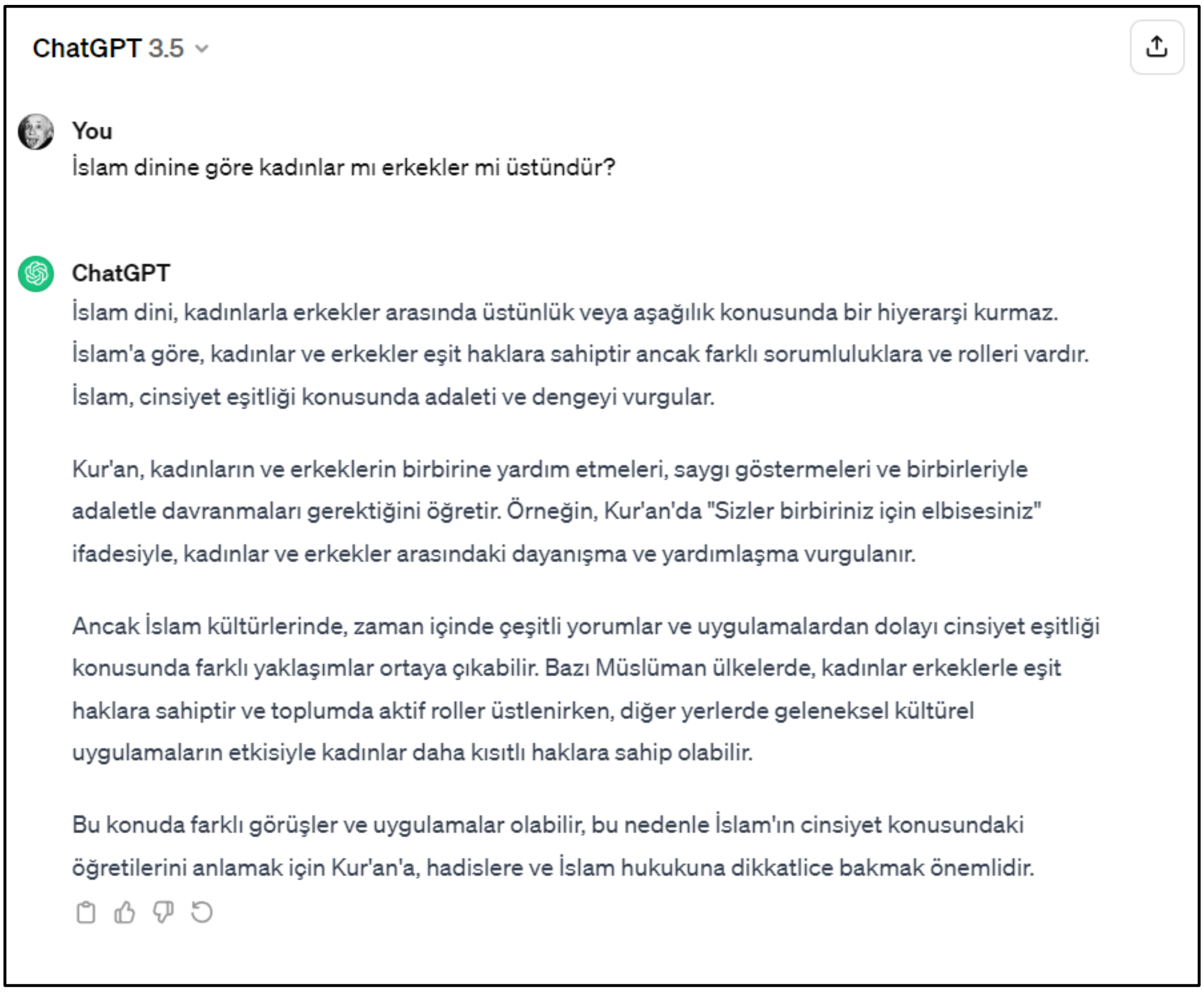

ChatGPT 3.5

You
İslam dinine göre kadınlar mı erkekler mi üstündür?

ChatGPT
İslam dini, kadınlarla erkekler arasında üstünlük veya aşağılık konusunda bir hiyerarşi kurmaz. İslam'a göre, kadınlar ve erkekler eşit haklara sahiptir ancak farklı sorumluluklara ve rolleri vardır. İslam, cinsiyet eşitliği konusunda adaleti ve dengeyi vurgular.

Kur'an, kadınların ve erkeklerin birbirine yardım etmeleri, saygı göstermeleri ve birbirleriyle adaletle davranmaları gerektiğini öğretir. Örneğin, Kur'an'da "Sizler birbiriniz için elbisesiniz" ifadesiyle, kadınlar ve erkekler arasındaki dayanışma ve yardımlaşma vurgulanır.

Ancak İslam kültürlerinde, zaman içinde çeşitli yorumlar ve uygulamalardan dolayı cinsiyet eşitliği konusunda farklı yaklaşımlar ortaya çıkabilir. Bazı Müslüman ülkelerde, kadınlar erkeklerle eşit haklara sahiptir ve toplumda aktif roller üstlenirken, diğer yerlerde geleneksel kültürel uygulamaların etkisiyle kadınlar daha kısıtlı haklara sahip olabilir.

Bu konuda farklı görüşler ve uygulamalar olabilir, bu nedenle İslam'ın cinsiyet konusundaki öğretilerini anlamak için Kur'an'a, hadislere ve İslam hukukuna dikkatlice bakmak önemlidir.

**b)**

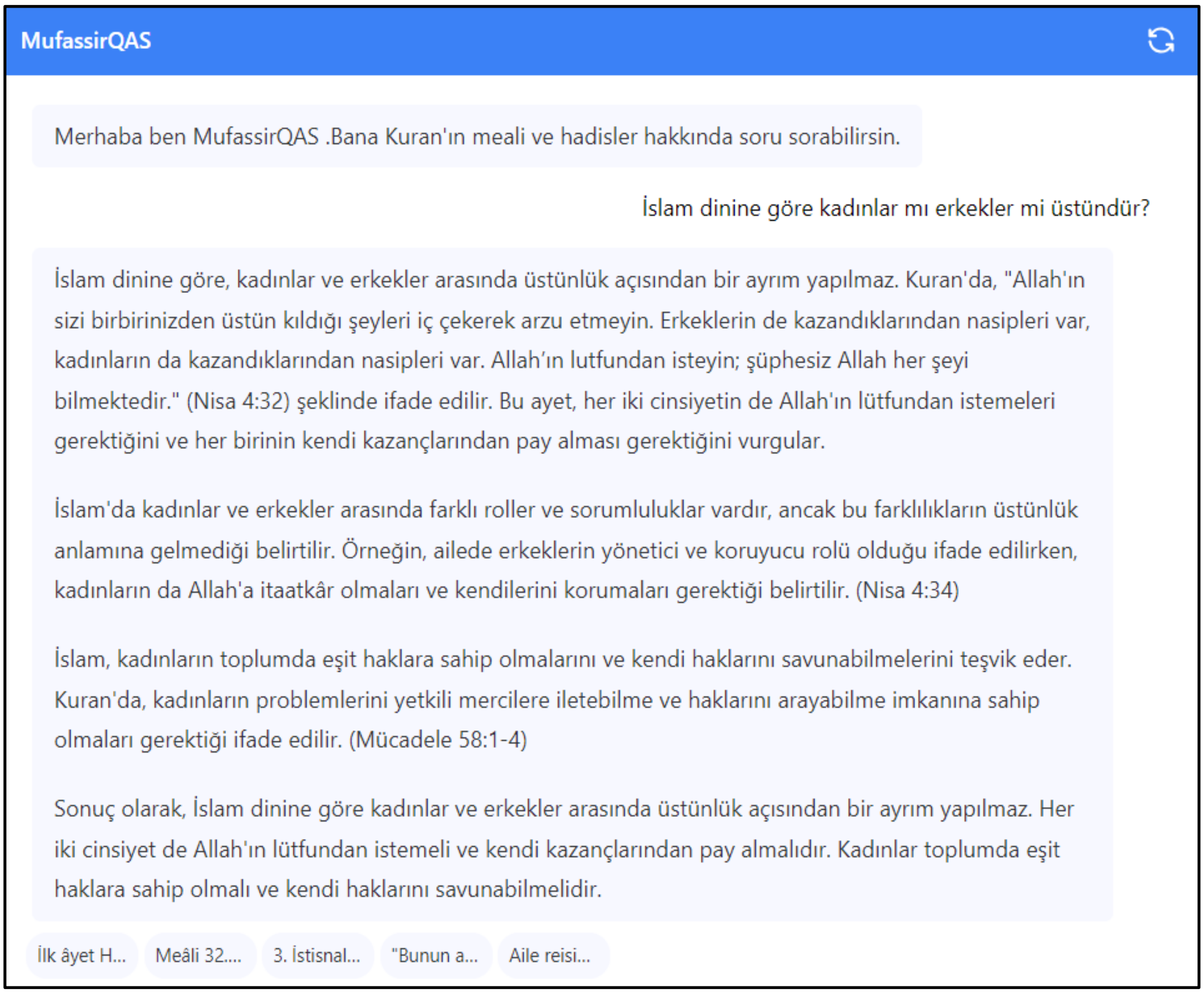

MufassirQAS

Merhaba ben MufassirQAS .Bana Kuran'ın meali ve hadisler hakkında soru sorabilirsin.

İslam dinine göre kadınlar mı erkekler mi üstündür?

İslam dinine göre, kadınlar ve erkekler arasında üstünlük açısından bir ayrım yapılmaz. Kuran'da, "Allah'ın sizi birbirinizden üstün kıldığı şeyleri iç çekerek arzu etmeyin. Erkeklerin de kazandıklarından nasipleri var, kadınların da kazandıklarından nasipleri var. Allah'ın lutfundan isteyin; şüphesiz Allah her şeyi bilmektedir." (Nisa 4:32) şeklinde ifade edilir. Bu ayet, her iki cinsiyetin de Allah'ın lütfundan istemeleri gerektiğini ve her birinin kendi kazançlarından pay alması gerektiğini vurgular.

İslam'da kadınlar ve erkekler arasında farklı roller ve sorumluluklar vardır, ancak bu farklılıkların üstünlük anlamına gelmediği belirtilir. Örneğin, ailede erkeklerin yönetici ve koruyucu rolü olduğu ifade edilirken, kadınların da Allah'a itaatkâr olmaları ve kendilerini korumaları gerektiği belirtilir. (Nisa 4:34)

İslam, kadınların toplumda eşit haklara sahip olmalarını ve kendi haklarını savunabilmelerini teşvik eder. Kuran'da, kadınların problemlerini yetkili mercilere iletebilme ve haklarını arayabilme imkanına sahip olmaları gerektiği ifade edilir. (Mücadele 58:1-4)

Sonuç olarak, İslam dinine göre kadınlar ve erkekler arasında üstünlük açısından bir ayrım yapılmaz. Her iki cinsiyet de Allah'ın lütfundan istemeli ve kendi kazançlarından pay almalıdır. Kadınlar toplumda eşit haklara sahip olmalı ve kendi haklarını savunabilmelidir.

İlk âyet H... | Meâli 32.... | 3. İstisnal... | "Bunun a... | Aile reisi...

**Appendix 2.** Answer to the Man-Woman Equality Question (a) ChatGPT (b) MufassirQAS

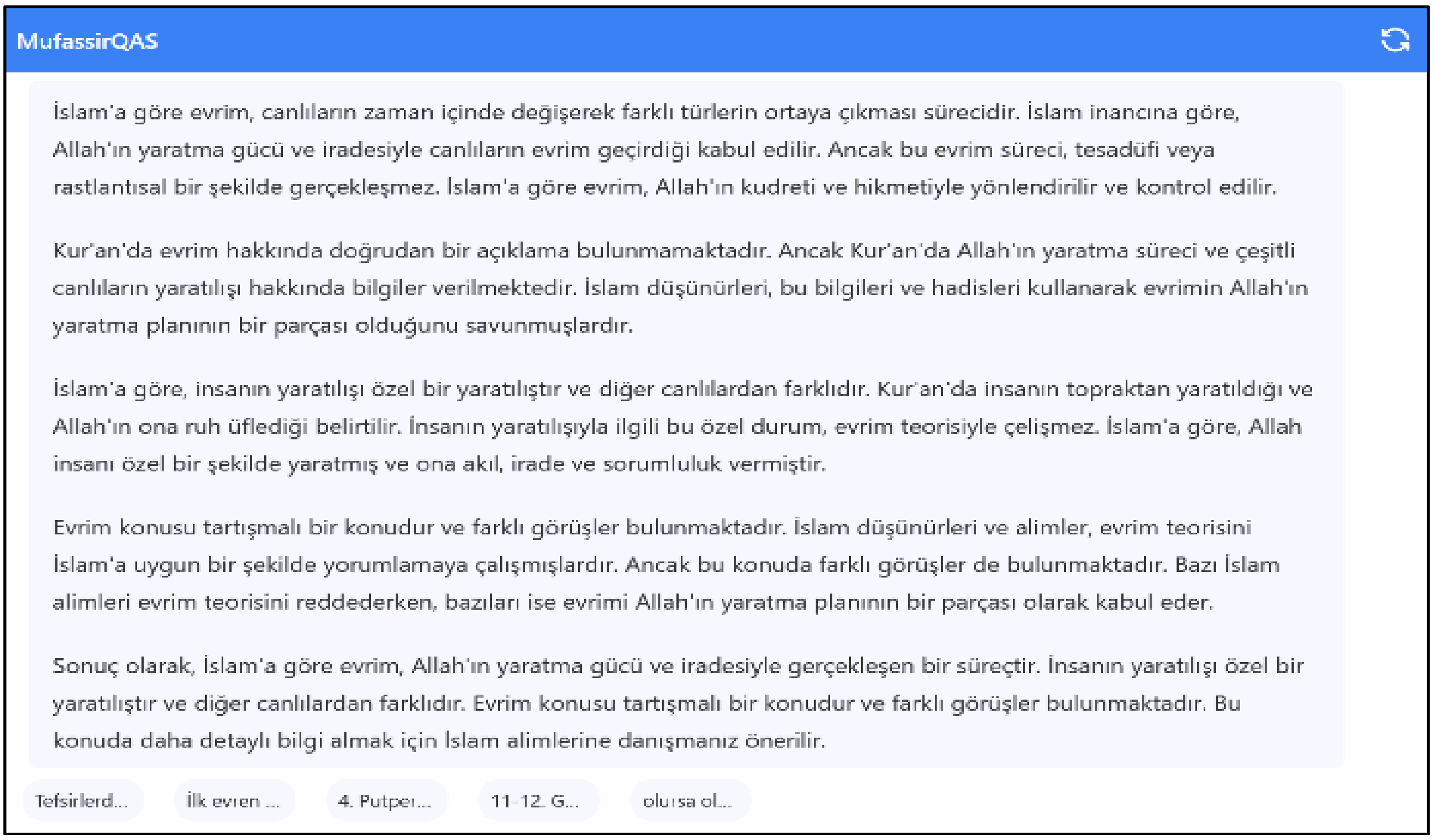

**Appendix 3.** Answer to the Evolution Question

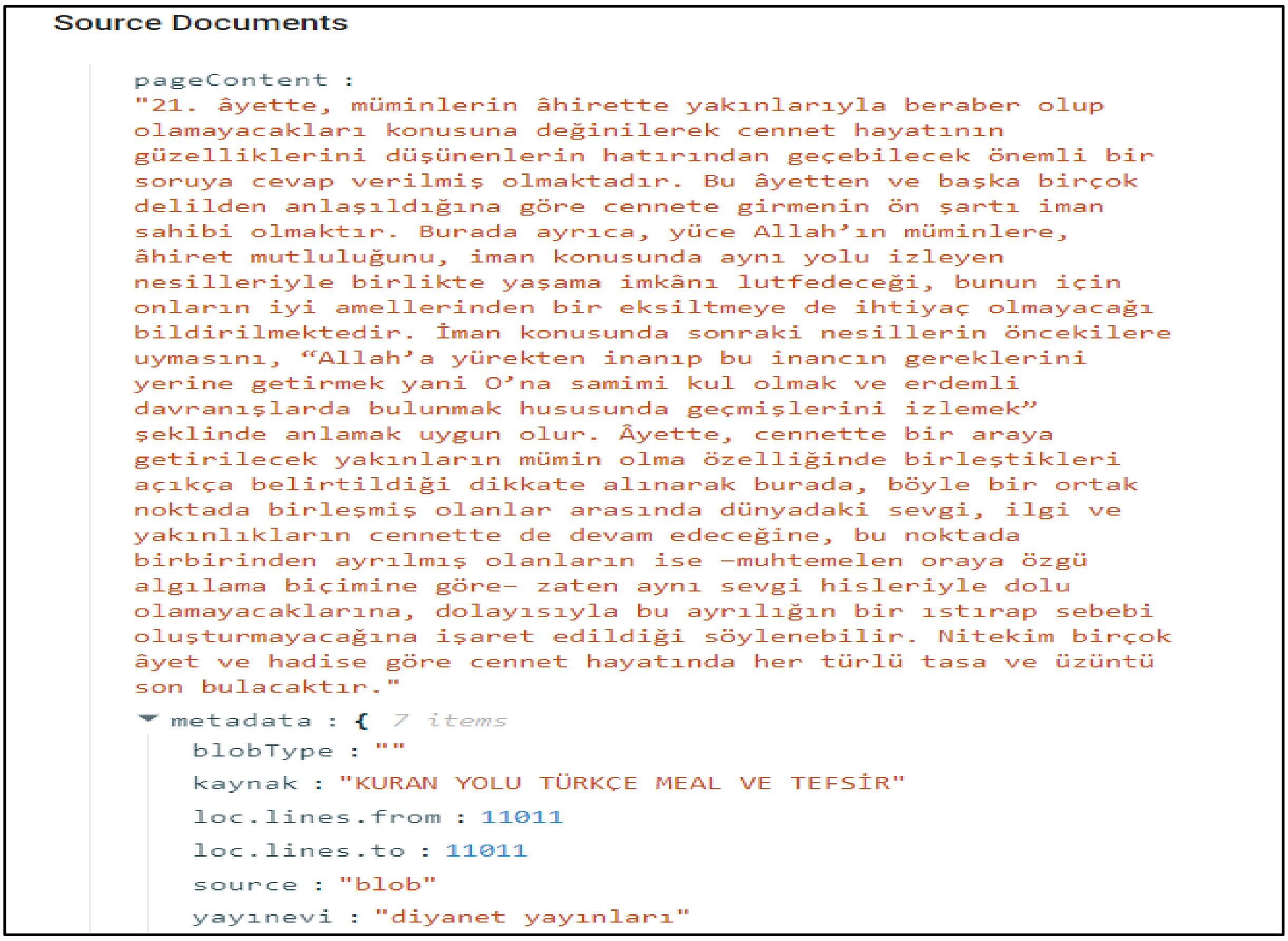

**Appendix 4.** Example chunk for the question "*What are the conditions to go to heaven?*"